\documentclass{esann}
\usepackage{graphicx}
\usepackage{wrapfig}
\usepackage{tikz}
\usepackage{subcaption}

\usepackage{amsmath,amsfonts,bm}
\usepackage[capitalise]{cleveref} 

\DeclareMathOperator*{\argmax}{arg\,max}
\DeclareMathOperator*{\argmin}{arg\,min}
\DeclareMathOperator*{\E}{\mathbb{E}} 
\newcommand{\norm}[1]{\left \lVert #1 \right \rVert}
\newcommand{\frnorm}[1]{\norm{#1}_\mathit{F}}

\newcommand{\Alpha}{\mathcal{A}}
\newcommand{\f}{f_\theta} 
\newcommand{\networksmanif}{\mathcal{F}_{\Alpha}}
\newcommand{\Loss}{\mathcal{L}} 
\newcommand{\dth}{\ensuremath{\delta\theta}} 
\newcommand{\given}{\;|\;}

\newcommand{\DU}{v_\nabla} 
\newcommand{\PU}{v^*}
\newcommand{\Vorth}{v_\perp} 
\newcommand{\EB}{\Psi} 

\newcommand{\optset}{\textit{train-opt}}
\newcommand{\lsset}{\textit{train-ls}}
\newcommand{\grset}{\textit{train-gr}}

\newcommand{\nicedisplay}[1]{#1}
\renewcommand{\nicedisplay}[1]{}

\newcommand{\compactdisplay}[1]{#1}

\title{

Growth strategies for arbitrary DAG neural architectures

}
\author{Stella Douka, Manon Verbockhaven, Théo Rudkiewicz, Stéphane Rivaud,\\
François P. Landes, Sylvain Chevallier, Guillaume Charpiat
\vspace{.3cm}\\
Inria TAU team, LISN, Université Paris-Saclay, Orsay, France}
\date{October 2024}

\begin{document}

\maketitle

\begin{abstract}
    Deep learning has shown impressive results, obtained at the cost of training huge neural networks. However, the larger the architecture, the higher the computational, financial, and environmental costs during training and inference. 
    We aim at reducing both training and inference durations. We focus on Neural Architecture Growth, which can increase the size of a small model when needed, directly during training using information from the backpropagation. We expand existing work and freely grow neural networks in the form of any Directed Acyclic Graph. 
    We design strategies that reduce excessive computations and steer network growth toward more parameter-efficient architectures.
\end{abstract}

\section{Introduction} 

\vspace{-1mm}
\begin{wraptable}{o}{0.\textwidth}
    \begin{tabular}{c|cc}
        Method & GPU days & kWh \\
        \hline
        Firefly & 1.5 & 9 \\
        NORTH & 0.4 & 2.4 \\
        ENAS & 0.45 & 2.7 \\
        DARTS & 1.5 & 9 \\
    \end{tabular}
    \caption{GPU power consumption \textbf{estimation} on CIFAR-10, assuming 250W power draw.
    }
    \label{tab:power_estimations}
\end{wraptable}

A common practice to train a deep architecture on a novel problem is to rely on over-parametrization -- meaning overly wide and deep networks -- as it facilitates optimization and yields better results. 
While it is possible to start with small models that are faster to train, they often lack expressivity and bear optimization issues. 
Hence, most literature focuses on training large neural networks and then using pruning, distillation, or compression to reduce energy consumption in the inference phase.
This includes training the large models and~fine-tuning them, requiring a tremendous amount of computational power and training time.
On the contrary, Neural Architecture Search methods (NAS) usually train multiple architectures from a finite set and choose the one that performs the best, usually with a trade-off between accuracy and cost. This is extremely resource-consuming and even Differential Architecture Search requires 1.5 GPU days to train on CIFAR-10 \cite{liu2018darts} (see Table \ref{tab:power_estimations}).
This is where Neural Architecture Growth comes at hand. The idea is to start with the simplest possible neural network and grow it by adding neurons in existing layers or adding entirely new layers, according to the information brought by the backpropagation. Such information can indeed be used to go beyond the usual limitations in small network training, to tackle potential optimization and expressivity issues.
The GradMax approach \cite{gradmax2022}, requires initializing all new input weights to zero, thus preserving the function's output. 
In NORTH \cite{maile2022when}, one measures the redundancy of the network as the orthogonality between post-activations. 
In Firefly \cite{firefly2020}, there is a choice between splitting existing neurons or creating new ones, which also includes adding new layers. All changes are kept local and one decides where to grow by solving the steepest-descent optimization problem. 
In the approach of \cite{Verbockhaven2024growing}, some of us introduce the notion of expressivity bottleneck to solve optimization issues in a sequential architecture by increasing its layers' width during training.
We refer to \cite{boumendil2024device} for a more thorough review of methods that help improve energy efficiency.

In this paper, we extend the work of \cite{Verbockhaven2024growing} by adding new layers on the fly, thus being able, for the first time, to grow neural networks in the form of any Directed Acyclic Graph. We test different strategies to grow the network efficiently and reduce energy~costs.


\section{Methodology} 

\newcommand{\TA}{\mathcal{T}_\Alpha}
\newcommand{\vvv}{v}
\newcommand{\FA}{\networksmanif}

\paragraph{Expressivity bottleneck.}
\begin{wrapfigure}{r}{0.4\textwidth}
\vspace{-0.6cm}
\input{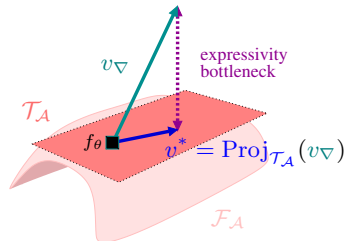}
\caption{Expressivity Bottleneck}
\label{fig:manifold}
\vspace{-0.3cm}
\end{wrapfigure}

In Figure \ref{fig:manifold} we present the concept of \emph{expressivity bottleneck} \cite{Verbockhaven2024growing}. 
We define the manifold $\networksmanif := \{ \f \given \theta \in \Theta_\Alpha \}$ as the functional space parameterized by a neural network, that is, the set of all possible functions one can represent by instantiating parameters of a fixed architecture $\Alpha$.
The tangent space at $\f$, namely  
$\mathcal{T}_\Alpha := \left\{ \frac{\partial \f}{\partial \theta} \dth \given \text{s.t. } \dth \in \Theta_\Alpha \right\}$,
consists of all the possible functions one can reach on the current manifold $\networksmanif$ using small parameter updates, \textit{e.g.} by gradient descent.
Now, let us denote by $\DU$ the desired update for the function $\f$ when we are not constrained by the current architecture:
\nicedisplay{
    \begin{equation}
        \begin{gathered}
        \DU(x) \;:=\; -\nabla_{f_{\theta}(x)}\Loss(f_{\theta}(x),\, y(x))
        \;:=\; -\nabla_a \Loss(a,y(x))\big|_{a = \f(x)}   \;\; .
        \end{gathered}
    \label{eq:desired_update}
    \end{equation}
}
\compactdisplay{
    $\DU(x) \;:=\; -\nabla_{f_{\theta}(x)}\Loss(f_{\theta}(x),\, y(x))
    \;:=\; -\nabla_a \Loss(a,y(x))\big|_{a = \f(x)}   \;\; .$
}
This is the functional gradient, \textit{i.e.}~the gradient of the loss w.r.t.~the output of the network.
The best update we can perform with the current architecture~$\Alpha$ is the projection $\PU$ of that desired update onto the tangent space $\mathcal{T}_\Alpha$.
As a result, the residual that should be completed by extending the network is:
\nicedisplay{
    \begin{equation}
        \Vorth := \DU - \PU
        \end{equation}
    \vspace{-0.3cm}
    \begin{equation}
        \text{ where } \qquad \PU \;:=\; \text{Proj}_{\mathcal{T}_\Alpha}(\DU) \;:=\; \argmin_{v \in \mathcal{T}_{\mathcal{A}}}\E_{(x, y)\sim \mathcal{P}} \left[\frnorm{\DU(x) - v(x) }^2 \right]
    \end{equation}
}
\compactdisplay{
    $\Vorth := \DU - \PU$, where
    $\PU \;:=\; \text{Proj}_{\mathcal{T}_\Alpha}(\DU) \;:=\; \argmin_{v \in \mathcal{T}_{\mathcal{A}}}\E_{(x, y)\sim \mathcal{P}} \left[\frnorm{\DU(x) - v(x) }^2 \right]$
}
and the residual's norm $\EB := \norm{\Vorth}$
is named the \emph{expressivity bottleneck} of the architecture.



One can add more neurons to a hidden state to mitigate the expressivity bottleneck at a given layer, thus growing the network. 
For further details, we refer the reader to the original paper.

\paragraph{Growing an arbitrary DAG.}
The contribution of this paper consists in the extension of the work by \cite{Verbockhaven2024growing} to non-sequential networks, in the form of any Directed Acyclic Graph (DAG) of fully connected layers. The graph in Figure \ref{fig:dag} shows an example of a non-sequential network where every edge represents a fully connected layer and each node represents a hidden state (or addition thereof). We optimize the new weights $\alpha,\omega$ so as to decrease the expressivity bottleneck:
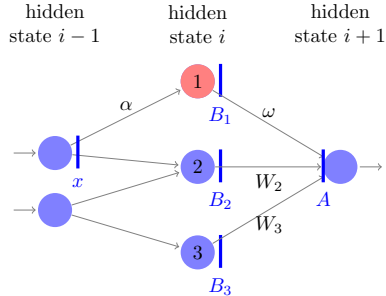
\begin{wrapfigure}{r}{0.45\textwidth}
    \resizebox{.45\textwidth}{!}{\def\layersep{2.5cm}

\begin{tikzpicture}[shorten >=1pt,->,draw=black!50, node distance=\layersep]
    \tikzstyle{every pin edge}=[<-,shorten <=1pt]
    \tikzstyle{neuron}=[circle,fill=black!25,minimum size=17pt,inner sep=0pt]
    \tikzstyle{input neuron}=[neuron, fill=green!50];
    \tikzstyle{new neuron}=[neuron, fill=red!50];
    \tikzstyle{hidden neuron}=[neuron, fill=blue!50];
    \tikzstyle{annot} = [text width=5em, text centered]

    \foreach \name / \y in {1,2}
        \node[hidden neuron, pin=left:] (L1-\name) at (0,-\y) {};

    \foreach \name / \y in {1,...,3}
    {
        \node[hidden neuron] (L2-\name) at (\layersep,50-\y*1.5 cm) {\y};
        \draw[-, blue, ultra thick] ([xshift=4mm]L2-\name.north) -- ([xshift=4mm]L2-\name.south) node[below] {$B_\name$};
    }
    
    \node[new neuron] (N) at (L2-1) {1};
    
    \node[hidden neuron,pin={[pin edge={->}]right:}, right of=L2-2] (L3) {};

    \path (L1-1) edge (L2-1);
    \path (L1-1) edge (L2-2);
    \foreach \dest in {2,3}
        \path (L1-2) edge (L2-\dest);

    \foreach \source in {1,...,3}
        \path (L2-\source) edge (L3);

    \node[annot,above of=L2-1, node distance=1cm] (hl) {hidden state $i$};
    \node[annot,left of=hl] {hidden state $i-1$};
    \node[annot,right of=hl] {hidden state $i+1$};

    \draw[-, blue, ultra thick] ([xshift=-3mm]L3.north) -- ([xshift=-3mm]L3.south) node[below] {$A$};
    \draw[-, blue, ultra thick] ([xshift=4mm]L1-1.north) -- ([xshift=4mm]L1-1.south) node[below] {$x$};

    \path (L1-1) -- (N) node[midway, above] (a) {$\alpha$};
    \path (N) -- (L3) node[midway, above] (o) {$\omega$};

    \foreach \name in {2,3}
    {
        \path (L2-\name) -- (L3) node[midway, below] (W-\name) {$W_\name$};
    }

\end{tikzpicture}}
    \caption{Example of DAG Network. \\We assume a new node added in color red with new weights $\alpha$ and $\omega$. We define pre-activities as A and post-activities as B or $x$.}
    \label{fig:dag}
    \vspace{-0.3cm}
\end{wrapfigure}
\nicedisplay{
    \begin{equation}
        \alpha^*\!,\,\omega^* = 
        \argmin_{\alpha,\, \omega} \| \,\omega \,\sigma(\alpha \cdot x) - \Vorth  \| 
    \label{eq:optimization_a}
    \end{equation}
}
\compactdisplay{
    $\alpha^*\!,\,\omega^* = 
        \argmin_{\alpha,\, \omega} \| \,\omega \,\sigma(\alpha \cdot x) - \Vorth  \| $.
}
With the current setting, we can create a network starting from an empty graph, rather than needing to choose a starting point. 
At each growth step, we have the option to add a direct edge (1 layer), add a new node (with 2 edges, \textit{i.e.}~2 new layers), or increase the size of an existing node by adding new neurons to its input and output layers (increase width). Expanding a node or adding a new one is the same process, as we only need to specify the input and output edges of the new neurons to be added. The peculiarity of this case lies in the fact that the best possible updates $\PU(x)$ of a specific node should take into account at least all the parameters contributing directly to this node. For reference, in Figure~\ref{fig:dag}, when calculating $\PU(x)$ at the hidden state $i+1$, we consider the pre-existing weights $W_2$ and $W_3$.


We split our training dataset into 3 equal parts named \textbf{\textit{train-opt}}, \textbf{\textit{train-ls}} and \textbf{\textit{train-gr}}.
At each growth step, we consider all possible network expansions.
For every possibility we optimize the new candidate neurons' weights
using \optset{}.
Then, for each new possible direction, we correct its amplitude by minimizing the loss on \lsset{}.
Finally, we keep only the best possible expansion, according to an estimate of the loss on \grset{}, and discard all the others.
We then train this newly expanded architecture with SGD, using the concatenation of \textit{train-opt} and \textit{train-ls}, which we refer to as \textbf{\textit{inter-train}}.

\paragraph{Strategies for Growth.}
The \textbf{\textit{whole search space}} described above is a greedy strategy, where we let the network grow freely based on the \textit{train-gr} loss alone.
The search space inflates very fast with the growth steps, together with the associated GPU energy consumption.
The \textbf{\textit{bottleneck restricted space}} strategy attempts to reduce this space by restricting the available network expansions. In this strategy, we find the node with maximum expressivity bottleneck $A^* = \argmax_A \EB_A$ and evaluate only the expansions that contribute to this pre-activity, that is, expanding or adding new layers that output to $A^*$ or expanding the node $A^*$ itself. This way we greatly reduce the search space and thus the search time and GPU energy consumption. 
In a third strategy, we aim at a trade-off between performance and complexity, within the bottleneck restricted search space. We use the Bayesian Information Criterion as $\text{BIC} = k \log(n) - 2 \log(\Loss)$, where $k$ is the number of parameters and $n$ is the sample size.
This strategy is named \textbf{\textit{BIC + restricted space}}. 
To compare these strategies, we consider an ideal situation where we already know the perfect architecture for the task, that is a Teacher's architecture, thought of as an oracle.

\section{Experiments and Results} 
\paragraph{Proof of concept.}

\begin{figure}
  \begin{minipage}[c]{0.75\textwidth}
    \includegraphics[width=\textwidth]{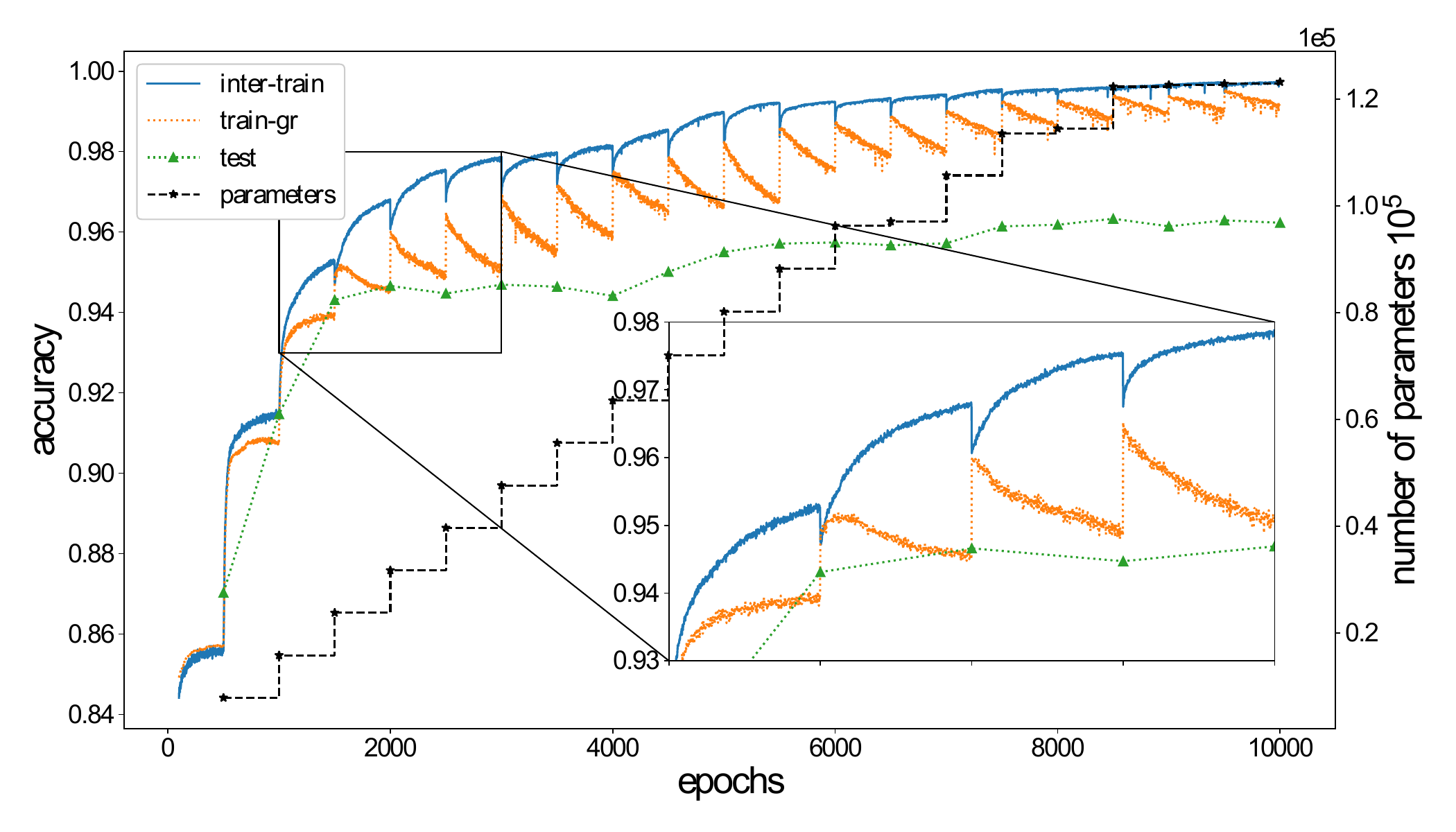}
  \end{minipage} 
  \hfill
  \begin{minipage}[c]{0.2\textwidth}
    \caption{
       Neural Architecture Growth results on MNIST using arbitrary DAG networks. We grow the architecture every 500 epochs.
    } \label{fig:mnist_results}
  \end{minipage}
  \vspace{-0.6cm}
\end{figure}

\begin{wrapfigure}{r}{0.4\textwidth}
\vspace{-0.3cm} 
    \centering
    \includegraphics[width=\linewidth]{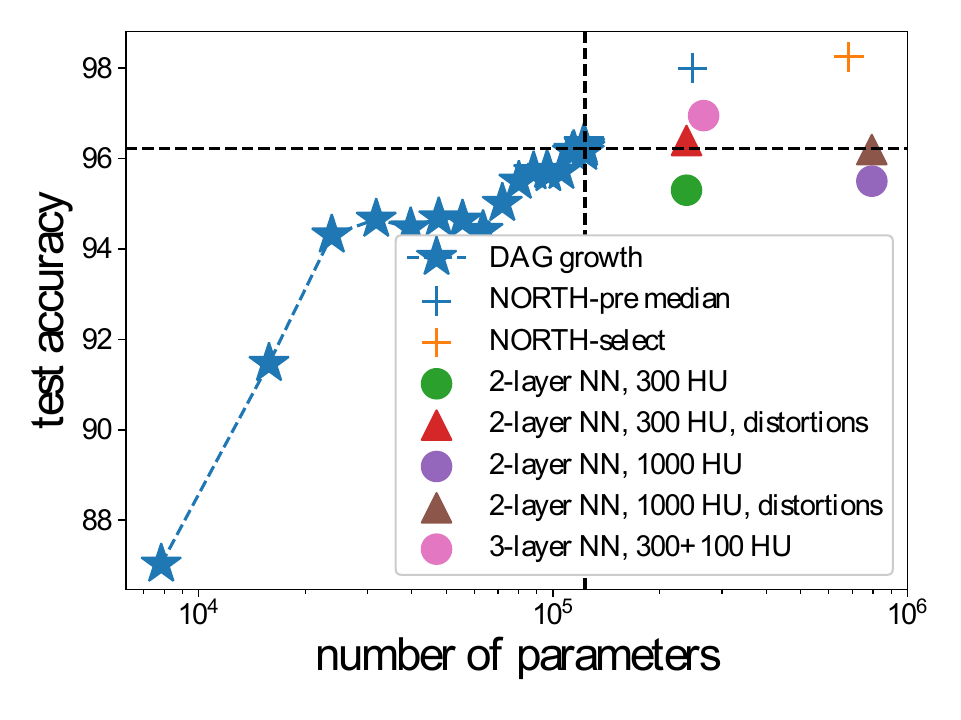}
    \vspace{-0.5cm}
    \caption{Baselines on MNIST for test accuracy and number of parameters.}
    \label{fig:mnist_baselines}
\vspace{-0.3cm} 
\end{wrapfigure}

In~\cite{Verbockhaven2024growing}, we evaluated growing networks on CIFAR-100 with sequential architectures. This study considers further experiments with DAG networks. As we have implemented only fully-connected layers so far, we are conducting a first evaluation on MNIST.
To the best of our knowledge, there are no published results of NAS methods on MNIST except for \cite{maile2022when}. For this reason, we use the results of \cite{lecun1998mnist} with fully-connected layers as our baselines. 
We use the \emph{whole search space} strategy and at each growth step, we increase the size of the architecture by 10 neurons. We perform intermediate training between growth steps for 500 epochs and evaluate on the test set.
We notice in Figure \ref{fig:mnist_results} that this intermediate training pushes parameters towards overfitting, but immediately after the architecture grows, the overfit gap between the \textit{inter-train} and \textit{train-gr} sets is reduced and the network finds itself in a more advantageous position, so it can continue learning.
In general, by growing we manage to escape potential local minima that force the training accuracy to converge and we gain significantly more accuracy on \textit{train-gr}. The test accuracy sits just below, since we also overfit on \textit{train-gr} after a few steps, as it is used for the expansion selection. Nevertheless, the gain in test performance is slow but significant. 
The experiment was run for 20 growth steps, requiring a little less than 7 GPU hours (0.29 GPU days).
The final architecture achieves an \textit{inter-train} accuracy of 99.7\% and a test accuracy of 96.2\%. We could keep growing the architecture for more steps but the improvement in accuracy is not significant and the drain on power consumption and additional complexity are not worth the added efficiency. In Figure \ref{fig:mnist_baselines} we see that we do not achieve state-of-the-art test accuracy but our method is extremely competitive in terms of model complexity and is thus cost-efficient.
We expect that strategies such as those presented in the next section may further improve this trade-off.

\paragraph{Growth Strategies.}

\begin{figure}
    \vspace{-0.3cm}
    \centering
    \begin{subfigure}{0.48\textwidth}
        \includegraphics[width=\textwidth]{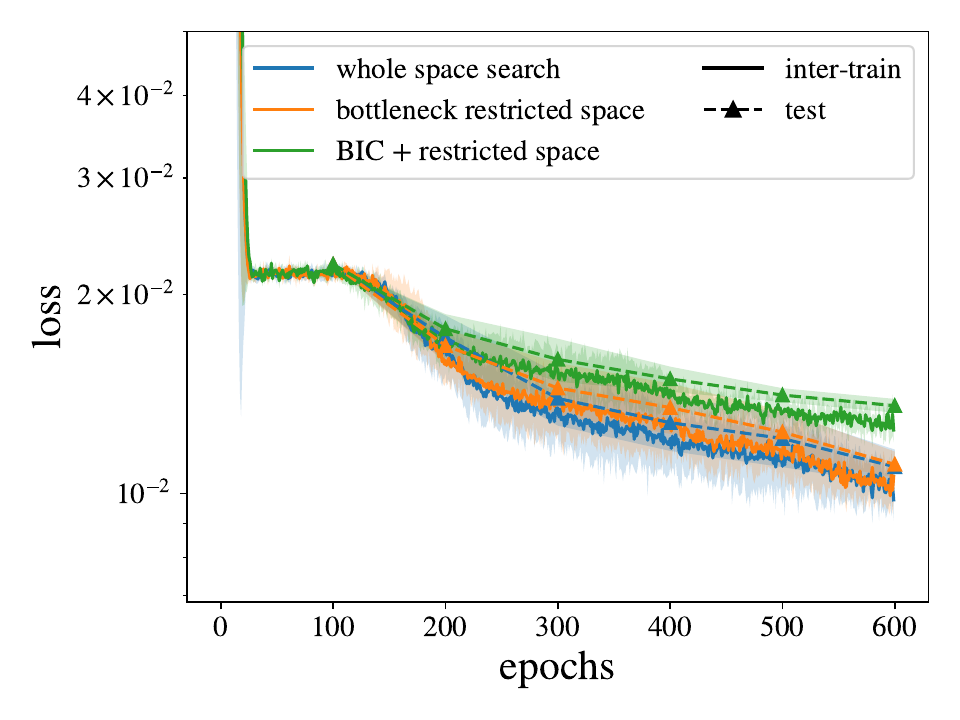}
    \end{subfigure}
    \hfill
    \begin{subfigure}{0.48\textwidth}
        \includegraphics[width=\textwidth]{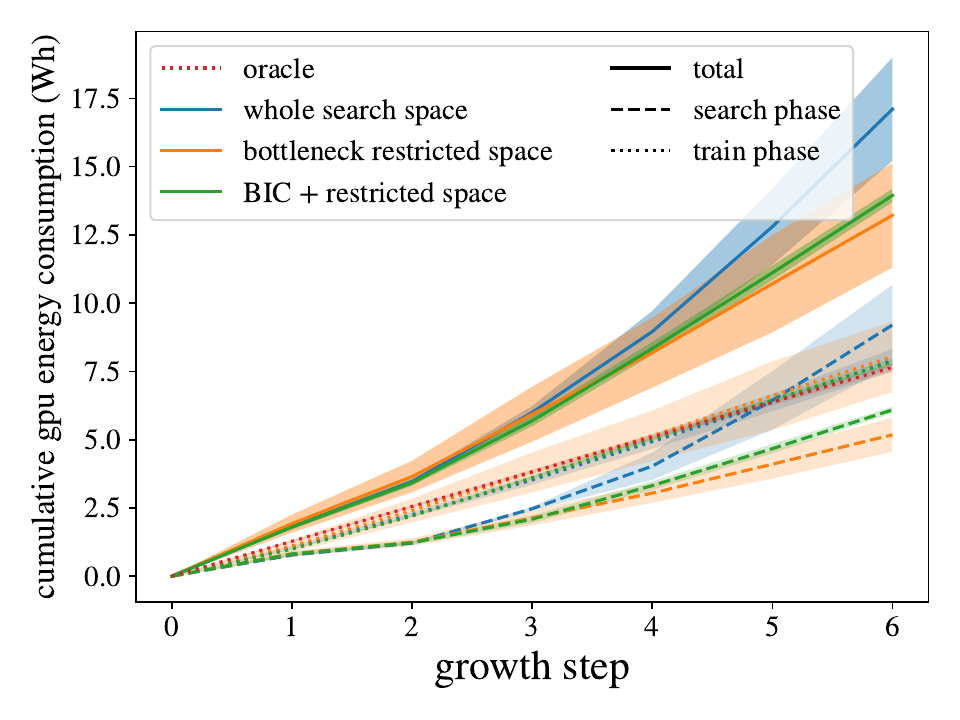}
    \end{subfigure}
    \vspace{-0.2cm}
    \caption{(a) Teacher-Student loss performance. The highlighted area represents the standard deviation over 6 runs. (b) Teacher-Student cumulative GPU energy consumption in Wh. Oracle: cost to train an architecture identical to the Teacher for the same number of epochs. The highlighted area represents the 85\% IQR over 6 runs.}
    \label{fig:exp_teacher}
    \vspace{-0.3cm}
\end{figure}

To compare strategies in a general framework, we construct a Teacher-Student experiment.
We randomly initialize a Teacher network for a regression task, with an input size of 20, two hidden states of size 50, a direct connection from the input to the second hidden state, and selu activations, for a total of 4701 parameters. 
We can then generate input samples from a uniform distribution and ask this Teacher for labels, to create our train and test datasets. 
Predicting this output is non-trivial as the intrinsic dimension is the same as the embedding dimension.
Indeed, the random initialization of the Teacher parameters and the independent sampling of the input features create many degrees of freedom.
We want to compare how our three strategies perform at growing a Student with arbitrary DAG, from scratch, to imitate the Teacher.
At each growth step, we can add 10 neurons and we perform intermediate training for 100 epochs.
The experiments are shown in Figure \ref{fig:exp_teacher}. 
We notice a temporary slight drop in performance when we restrict the search space, but it is not a significant one and it disappears after a certain number of epochs.
However, GPU energy consumption is decreased by 23\% when restricting the search space, for the same loss. We note also that with this strategy, we only consume 70\% more energy than the oracle (the ideal scenario where the original Teacher architecture is known and trained from scratch), while here we also have to find the architecture.

Based on these results we estimate that with a grid-search NAS technique instead, we would need to train 7 different architectural structures for 100 epochs to roughly evaluate all the possible DAGs we can achieve for a network of 5 layers. Assuming they would all have the same number of neurons, chosen among a grid of 5, we would have to train and test at least 35 architectures, plus the best architecture fully once, for a total of 
$\approx 51$ Wh. We achieve a more granular result with only 25\% of the energy when restricting the search space. In retrospect, reducing the search space based on the bottleneck seems to perform very well in terms of efficiency and cost.

The use of BIC reduces the size of the resulting architecture, with an average of 723 parameters compared to 1479 for the \emph{bottleneck restricted space} and 1454 for the whole search space, but consumes more energy during the search phase than just reducing the space, although with a smaller variance. This is because it tends to choose architectures that create more options for the next search phases.
Studying variations on BIC, that may be able to achieve a better performance/architectural complexity trade-off, is left for future work.

\section{Conclusion and Future work} 
In this work, we grow neural architectures in the form of any DAG by adding new layers and direct connections on the fly during training.
Our work is based on \cite{Verbockhaven2024growing}, which introduced the notion of expressivity bottleneck to increase the width of existing layers in a pre-defined architecture structure. 
Our contribution is to create arbitrary non-sequential fully connected architectures starting from an empty graph, without any predefined structure. We show that our method is competitive in terms of the number of parameters, thus reducing inference time. We compare various strategies to grow an architecture and achieve lower complexity. We manage to reduce the overall training time and thus the GPU energy consumption compared to a grid search among architectures. 
The next line of research is to further improve our strategy
to fulfill an efficient trade-off between performance and complexity. 
We intend to further extend our work to introduce growable modules for convolutional layers and address scalability.

\paragraph{Acknowledgments}
This work was supported by grants ANR-22-CE33-0015-01 and ANR-17-CONV-0003 operated by LISN to S.Chevallier and by ANR-20-CE23-0025 operated by Inria to G.Charpiat. 
The European Union also funded this work under GA no.101135782 (MANOLO project).
The development, experiments, and research tools used in this paper consumed roughly 42.5kWh \cite{GA4HPC}, equivalent to 2.18kg of CO2eq in France or flying 105 miles in economy class.

\bibliographystyle{acm}
{\scriptsize
\bibliography{sources}
}
\end{document}